# Guided Self-attention: Find the Generalized Necessarily Distinct Vectors for Grain Size Grading

Fang Gao, Xuetao Li, Jiabao Wang, Shengheng Ma, Jun Yu

***Abstract*—With the development of steel materials, metallographic analysis has become increasingly important. Unfortunately, grain size analysis is a manual process that requires experts to evaluate metallographic photographs, which is unreliable and time-consuming. To resolve this problem, we propose a novel classification method based on deep learning, namely GSNets, a family of hybrid models which can effectively introduce guided self-attention for classifying grain size. Concretely, we build our models from three insights:(1) Introducing our novel guided self-attention module can assist the model in finding the generalized necessarily distinct vectors capable of retaining intricate relational connections and rich local feature information; (2) By improving the pixel-wise linear independence of the feature map, the highly condensed semantic representation will be captured by the model; (3) Our novel triple-stream merging module can significantly improve the generalization capability and efficiency of the model. Experiments show that our GSNet yields a classification accuracy of 90.1%, surpassing the state-of-the-art Swin Transformer V2 by 1.9% on the steel grain size dataset, which comprises 3,599 images with 14 grain size levels. Furthermore, we intuitively believe our approach is applicable to broader applications like object detection and semantic segmentation.**

## I. INTRODUCTION

GRAIN size determination is a critical step in ensuring steel quality. Traditional techniques of this procedure, however, are typically conducted manually, which is inefficient and unreliable for measuring the right grain size and size distribution, especially when the grains rises exponentially. With the fast progress of deep learning, computer vision has witnessed booming technological development in the past thirty years. Specifically, due to the ability of precise measurements and rapid analysis, deep learning technology is widely applied in almost every field of our lives. To improve the accuracy of grain size determination, it is advisable to allow a machine to automatically grade steel grain size levels using computer vision technology (see Fig.1).

This work was supported by the Guangxi Science and Technology Base and Talent Project under Grant 2020AC19253, CAAI-Huawei MindSpore Open Fund CAAIXSJLJJ-2021-016B, Anhui Province Key Research and Development Program 202104a05020007 and USTC Research Funds of the Double First-Class Initiative YD2350002001. *(Fang Gao and Xuetao Li are co-first authors.) (Corresponding author: Jun Yu.)*

Fang Gao, Xuetao Li and Jiabao Wang are with Guangxi Key Laboratory of Intelligent Control and Maintenance of Power Equipment, School of Electrical Engineering, Guangxi University, Nanning 530004, China (email: fgao@gxu.edu.cn, xtli312@163.com, 1912302010@st.gxu.edu.cn)

Shengheng Ma is with Guangxi China-Tek Blue Valley Semiconductor Technology Co., Ltd. (zkma@ctnuite.com)

Jun Yu is with Department of Automation, University of Science and Technology of China, Hefei 230027, China (email: harryjun@ustc.edu.cn).

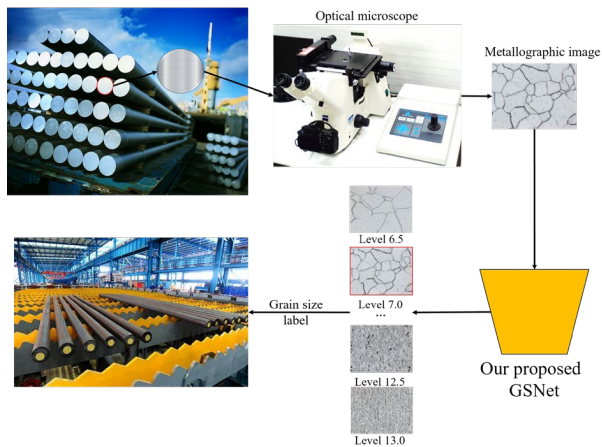

Fig. 1. Overview. First, we would take a slice from the steel that needed to be graded. Then the slice would be processed by the Zeiss Axiovert 200 Mat optical microscope to produce the metallographic image. The metallographic image would be fed into our proposed GSNet (Guided Self-attention Net) to obtain the grain size label for the steel grading.

Convolutional Neural Networks (CNNs) had dominated the field of computer vision for many years after AlexNet [1] was proposed. With the success of Transformers [2] in natural language processing [3], researchers attempted to introduce self-attention into computer vision. Swin Transformer [4] recently provided a hierarchical transformer backbone, outperforming previous state-of-the-art models in a wide range of vision tasks such as image classification, object detection, and object localization [5], thus showing huge potential for use in the vision domain.

Because Swin Transformer was built directly on the Vision Transformer (ViT) [6], it might fall behind CNNs in the low data regime, as ViT did [7]. Existing work on Swin Transformer mainly focuses on large datasets, with little work on small ones. Some works attempted to adapt receptive fields[4], [6], [8] or explicit convolutions [9], [10] in Transformer models, which might improve the performance by incorporating the inductive biases of CNNs into the self-attention mechanism. Those approaches mainly focused on a particular incorporated way, lacking a systematic understanding of the respective roles of convolution and self-attention when combined [11], [12]. Some SOTA networks such as MaxViT [13], and Swin Transformer V2 [5] are large-scale models requiring numerous parameters and extensive training on very large datasets. Achieving high accuracy often relies heavily on fine-tuning expertise and substantial computational resources [14], [15]. Unfortunately, fine-tuning tricks are usually undisclosed, and the cost of GPU or TPU servers is exorbitant. Consequently, many researchers resort to using pre-trained models to improve the performance of their models [11], [16], [17]. However, most pre-trained networks have limited performance on smaller datasets in specialized domains like industry, agriculture, electricity, and so on [18], [19]. These domains significantly impact our daily lives, but due to confidentiality concerns, images from these sectors are typically unavailable in public datasets. Thus, there arises a need for a network capable of robust generalization in specialized scenarios without relying on pre-trained models or fine-tuning expertise.

In this work, we propose GSNet, which can outperform MaxViT [13], ConvNext_CLIP [16], EVA-02 [17], CoAtNet [11], Swin Transformer V2 [5] without the need for a pre-trained model, fine-tuning skills, and extra datasets. Specifically, our study indicates the high-level feature map in CNNs tends to have higher pixel-wise linear independence, while Swin Transformer had a greater ability for capturing global information due to the shifting window scheme. As a result, when Swin Transformer is introduced with our guided self-attention, a substantially stronger representation with increased pixelwise linear independence may be obtained. It is noteworthy that our best GSNet version outperforms the SOTA model Swin Transformer V2 with the ImageNet-22K pretraining model by 1.9% in grain size grading accuracy even without pretraining. The main contributions of this paper can be summarized as follows.

1) We propose an encoder module to enhance the pixelwise linear independence of the feature map and thereby capture the highly condensed semantic representations;

2) We propose a guided self-attention module to guide the model to find the generalized necessarily distinct vectors, which can retain intricate relational connections and rich local feature information.;

3) We propose a triple-stream merging module to improve the generalization capability of the model by combining three feature-extracting strategies.

4) We propose the IAWCA (Improved Adaptive Weighted Channel Attention) module to dynamically give attention weight to different channels.

The rest of this paper is laid out as follows. Section II gives a summary of related works. Section III explains our model and the proposed architecture. Experiments and ablation studies are performed in Section IV. Section V concludes this paper.

## II. RELATED WORK

### A. Traditional Grain Size Grading Methods

The grain size determines many material qualities such as strength, creep, fatigue resistance, electrical, and magnetic properties. To explore the relationship between microstructure and characteristics in martensitic steels, the prior austenite grain size must be measured [20]. Currently, ASTM E112 [21] and ISO 643 [22] standards define three ways to determine the grain size: the comparison procedure, the intercept procedure, and the planimetric procedure.

The comparison procedure determined the grain size via the degree of similarity between the grain structure and the comparison chart of sizes, while the intercept and planimetric procedures were determined by assessing the number of grains contained within a defined test region. Those standard procedures were the principal methods for grain size grading. For the comparison procedure, when the operator performed multiple checks on the same specimen, the operator was predisposed to the first estimate. Therefore, different operators may give different comparison results. In contrast, the planimetric and intercept procedures could both provide grain size accuracy of 0.5 units and reproducibility of 0.25 grain size units, so the majority of measurement operators employed the planimetric and intercept procedures for grain size grading. Although the intercept and planimetric procedures were effective for estimating grain size, they needed more time due to picture pre-processing compared with the comparison procedure.

### B. Deep Learning Method

Although the CNNs had existed for several decades [23], it was not until the breakthrough of AlexNet that they started to dominate the field of computer vision. Then the structure of CNNs had become deeper [24], [25], [26], and wider [27]. These advances have propelled the deep learning wave in computer vision even further. Besides these architectural advancements, substantial work had been done to enhance the connection between convolution layers, such as ResNet [28], whose ResNet blocks were popular in nowadays large-scale CNNs. Following the success of ResNet, DenseNet proposed a dense connection technique that could reuse the local features while avoiding information loss due to down-sampling. In our proposed GSNet, Dense blocks are used as convolution building blocks.

The development of transformer-based models in computer vision has advanced significantly since the Transformer was proposed [29], [30]. The ViT pioneered the use of the Transformer architecture for image classification on non-overlapping medium-sized picture patches. It outperformed convolutional networks in terms of speed and accuracy in image classification [31]. Unlike ViT, which required large-scale training datasets (JFT-300M), DeiT [7] provided different training approaches that allowed ViT to function on the smaller ImageNet-1K dataset as well. The results of ViT on image classification were promising, but its low-resolution feature maps and quadratic computation complexity with input picture size precluded its usage as a general-purpose backbone network for dense vision applications. In contrast, Swin Transformer was linear and hierarchical, which enabled it to be quick and accurate, outperforming previous state-of-the-art classification models on ImageNet. In this work, Swin Transformer is chosen as the Transformer building block.

Computer vision and image processing have been applied in microstructure science [32], [33]. Zhang et al. [34] proposed a fuzzy logic-based technique for edge identification in metallographic pictures. To get more precise edges, Gajalakshmi et al. [35] used Canny edge recognition and support vector regression (SVR) to develop an image processing approach for determining the average grain size in a metallic microstructure. In terms of model capacity, SVR fell short of CNNs. Ma et al. [36] applied U-Net [37] for grain boundary identification in polycrystalline materials and a CNN for grain object tracking across slices to reconstruct the sample's 3D structure. Lee et al. [38] adopted ResNet and cross-stage partial networks (CSPNet) [39] to extract local descriptors from each image. However, the method [38] only discussed the performance on the datasets with four grain size levels, lacking further discussion for full-scaled grain size levels.

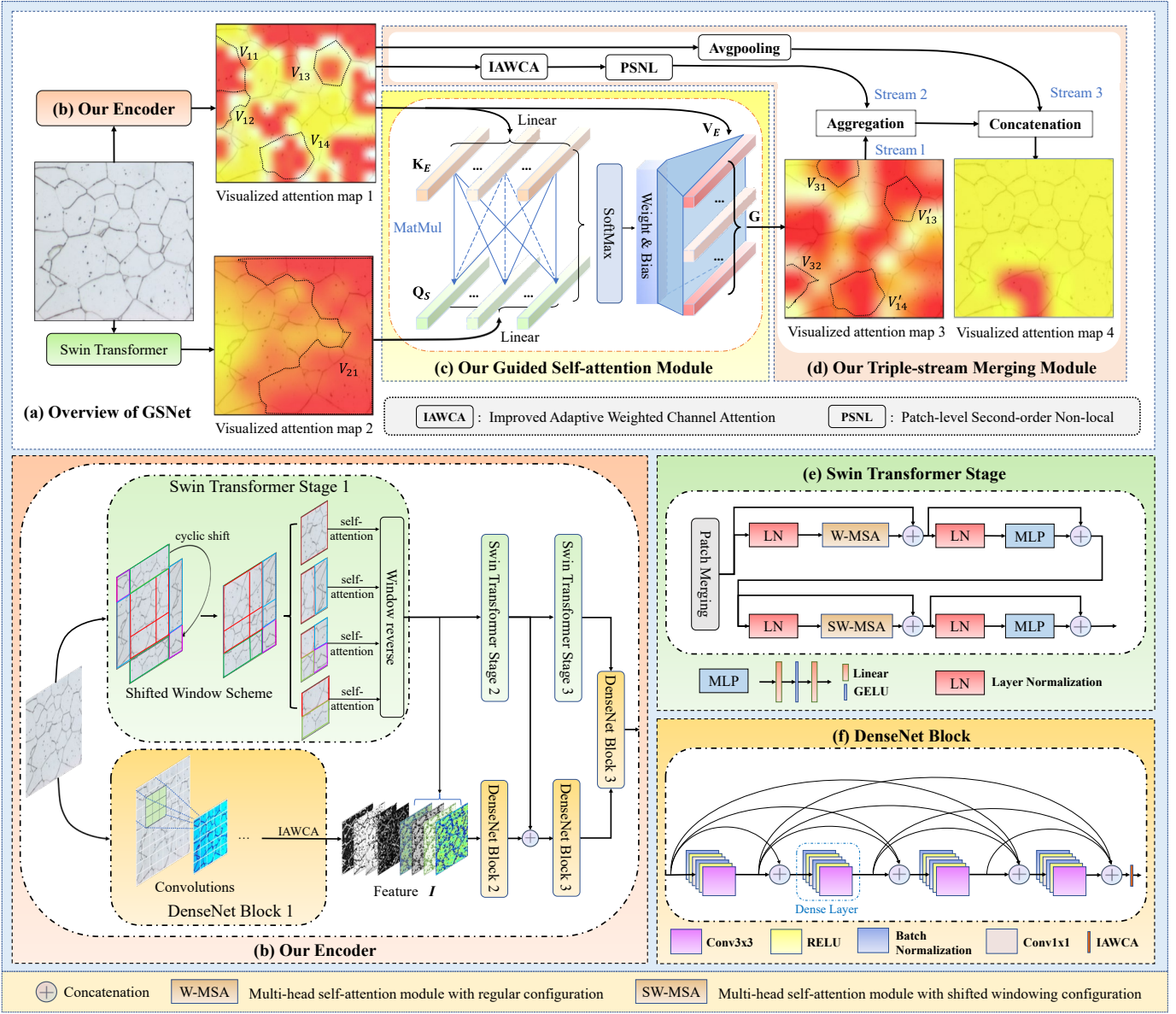

Fig. 2. (a) An overview of our proposed GSNet. It is comprised of (b) Our Encoder, (c) Our Guided Self-attention Module, and (d) Our Triple-stream Merging Module. (b) Our Encoder consists of (e) the Swin Transformer Stage and (f) the DenseNet Block. Firstly, the metallographic photographs will be processed by our encoder and produce a feature map that can efficiently and comprehensively merge global information (i.e., the steel grains $V_{11}$ and $V_{12}$, which can interact with each other even though they come from different regions of the feature map) and local information (i.e., the individual steel grains $V_{13}$ and $V_{13}$, which may represent the average grain size). Then, our guided self-attention module guides Swin Transformer, which can exploit global interaction (i.e., the red part $V_{21}$) effectively, to find out the generalized necessarily distinct vectors $\boldsymbol{G}$, which can not only learn the interactions between the grains (i.e., $V'_{13}$ and $V'_{14}$) that are far away, but also locate appropriate individual grains (i.e., $V_{31}$ and $V_{32}$) that can represent the average grain size for grain size grading. Finally, our triple-stream merging module merges three feature maps coming from three different mechanisms to locate an individual grain for grain size determination. Besides, the $\boldsymbol{K_E}$, $\boldsymbol{V_E}$ and $\boldsymbol{Q_S}$ are matrices mapped linearly from feature maps.

## III. MODEL

Our proposed model aims to "optimally" assist Transformer in finding the generalized necessarily distinct vectors. The overall architecture of the proposed method is shown in Fig. 2, consisting of our encoder module, our guided self-attention module, and our triple-stream merging module.

### A. Encoder Module

In Fig. 2(b), our encoder module effectively encodes images into highly condensed semantic representations. This differs from conventional transformer-based encoders by capturing feature maps containing greater pixel-wise linearly independent information, enhancing its efficacy in fitting downstream tasks. We borrow ideas of the shifted window from Swin Transformer since it could enhance the cross-window connection in the feature map, as well as maintaining a hierarchical design, which is advantageous for feature map aggregation. Specifically, each Swin Transformer stage contains patch merging layers and two successive Swin Transformer blocks (see. Fig. 3(a)). The patch merging layers are used to expand the reception field and channel size of the feature map. The Swin Transformer block1 and block2 are multi-head self-attention modules with regular (i.e., W-MSA shown in Fig. 3(b)) and shifted windowing (i.e., SW-MSA shown in Fig. 3(c)) configurations, respectively [4], respectively. When the input of the Swin Transformer stage is $H \times W \times C$ ($H \times W$ is the feature map size and $C$ is the channel dimension), the output will be $H/2 \times W/2 \times 2C$, indicating that the architecture of Swin Transformer can provide a hierarchical representation as CNNs.

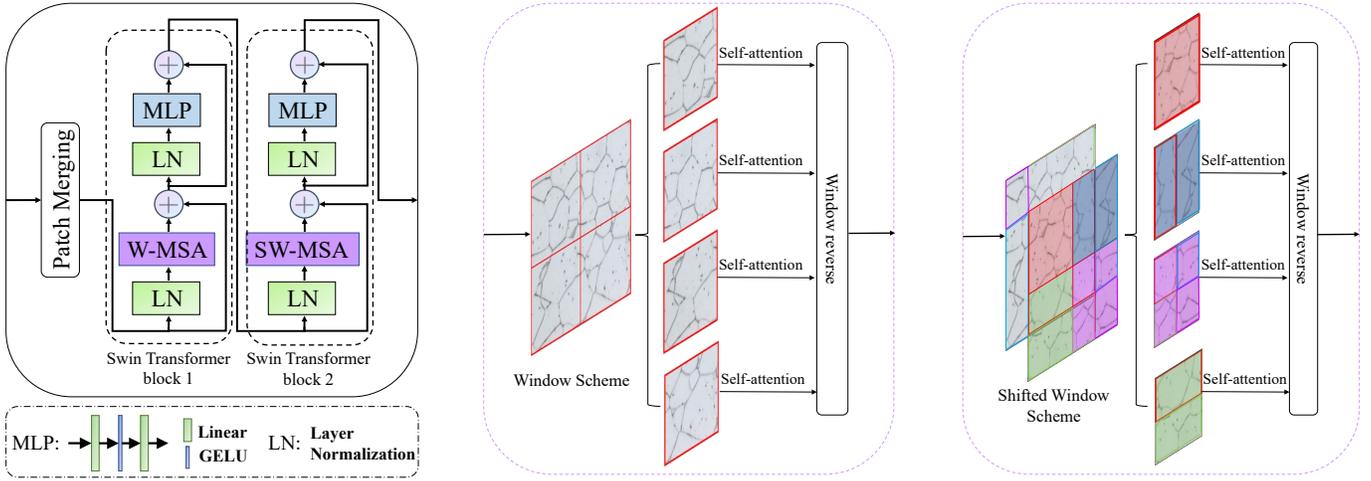

Fig. 3. (a) Structure of Swin Transformer stage; (b) W-MSA: Multi-head self-attention module with the regular configuration; (c) SW-MSA: Multi-head self-attention module with the shifted windowing configuration.

TABLE I
MODEL PERFORMANCE ON THE STEEL GRAIN SIZE DATASET. (#ACC (WITH/WITHOUT): THE Top-1 ACCURACY OF MODEL WITH OR WITHOUT LOADING THE PRETRAINED MODEL, #Params: PARAMETER SIZE, THE INPUT IMAGE SIZE IS 224×224)

| Model | #Acc (without) | #Acc (with) | #Params |
|---|---|---|---|
| Swin-S | 0.760 | 0.879 | 188M |
| Swin-B | 0.764 | 0.883 | 332M |
| DenseNet121 | 0.854 | 0.865 | 27M |
| DenseNet161 | 0.863 | 0.873 | 102M |
| DenseNet201 | 0.862 | 0.870 | 70M |
| ViT-B | 0.827 | 0.897 | 327M |
| ViT-L | 0.818 | 0.896 | 1.13G |
| ResNet50 | 0.766 | 0.866 | 90M |
| ResNet101 | 0.769 | 0.869 | 162M |
| ResNet152 | 0.785 | 0.864 | 222M |
| ResNeXt50 | 0.755 | 0.876 | 88M |
| ResNeXt101 | 0.789 | 0.874 | 332M |
| Inception_v3 | 0.842 | 0.867 | 83M |
| EfficientNet-b0 | 0.761 | 0.875 | 15M |

Prior to the introduction of the Swin Transformer, Convolutional Neural Networks (CNNs) were the dominant machine learning approach in computer vision. CNNs excelled at exploiting local features, often outperforming Transformer-based designs in this regard [11]. From this point of view, we have evaluated several CNN-based models as a baseline for steel grain size grading, including DenseNet, ResNet, ResNeXt, Inception [40], and EfficentNet [15]. Table I compares the accuracy of Transformer-based models such as Swin Transformer and ViT with those CNN-based models. As shown in Table I, without extra ImageNet-22K pre-training, the classification accuracy of Swin Transformer is lower than DenseNet121 on the steel grain size dataset [41], which contains 3,599 images with 14 grain size levels. What is more, DenseNet121 has the second smallest parameter size and the third highest accuracy (without loading the pretrained model) on the steel grain size dataset. For the consideration of the tradeoffs between parameters and accuracy, we choose the DenseNet Block (see Fig. 2(f)) of DenseNet121 [42] as a component of CNNs in our model.

To harness a broader spectrum of global information in our encoder architecture, we implement a shifted window scheme (see Fig. 2(b)), which facilitates interactions between distant image patches. Subsequently, by employing a multi-head self-attention mechanism, we gather comprehensive global representations (i.e., the red part $V_{21}$ in Visualized attention map 2 of Fig. 2(a)) from each recombined patch. Concurrently, we concatenate the abundant global information with local information processed through convolutions (see Feature $I$ in "Our Encoder" of Fig. 2(b)) in the channel dimension. Feature $I$ advances to the subsequent DenseNet Block, ensuring continued enhancement of pixel-wise linear independence within the feature map containing both local and global information. Moreover, our channel-wise attention mechanism, IAWCA, allocates adaptive weights across all channels, enabling more comprehensive feature interaction between different channels. This iterative refinement via our encoder module persists until the feature map enters our guided self-attention module.

In our encoder module, the Swin Transformer and DenseNet

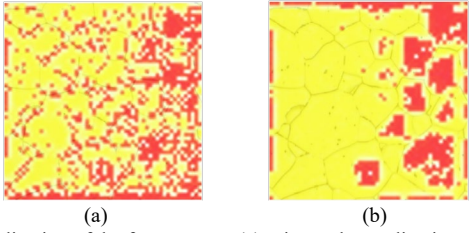

Fig. 4 Visualization of the feature map: (a) prior to the application of DenseNet Block 2, and (b) following the application of DenseNet Block 2.

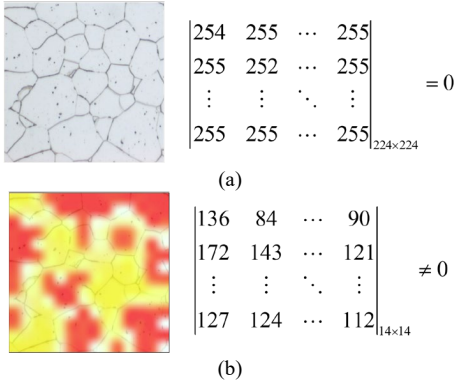

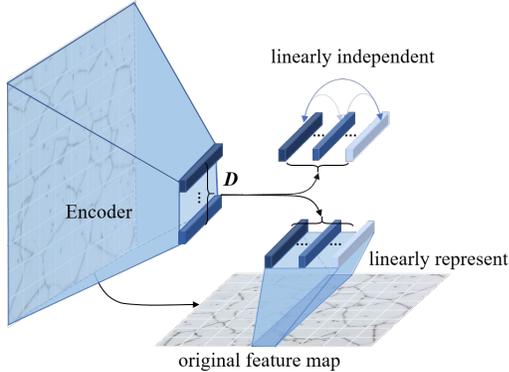

Fig. 5 Visualization and pixel matrix determinants of (a) the original picture, and (b) the encoder-processed feature map

Fig. 6. The illustration of the necessarily distinct vectors $\boldsymbol{D}$.

block are utilized as parallel branches to leverage their respective strengths in capturing different aspects of feature representations. Prior to concatenation, we align the dimensions of the features from both branches through a series of convolutional or linear layers to ensure that their feature dimensions are compatible. Subsequent to this concatenation, we introduce a new DenseNet block to merge features by convolution operations across different feature map channels. This option not only aids in smoothing any discrepancies between the feature maps but also enhances the composite feature map's representational capacity. As shown in Fig. 4(a), the feature map is inconsistent and chaotic after concatenation. However, after being processed by a DenseNet block, the inconsistent feature map is remapped to a new feature map constrained by the Cross-Entropy loss function. This remapping makes the feature map more consistent and better suited for identifying steel grains, which serves as a representative average size for steel grains (as illustrated in Fig. 4(b)).

When the determinant of a matrix is zero, it indicates that the columns of the matrix are linearly dependent. Conversely, a non-zero determinant signifies linear independence among the columns. To demonstrate the enhancement in pixel-wise linear independence afforded by our encoder architecture, we conducted a detailed analysis involving both visualization and quantitative assessment of the determinants of pixel matrices. We computed the determinants for both the original image and the encoder-processed feature map. As illustrated in Fig. 5, the feature map processed by our encoder module exhibits increased pixel-wise linear independence.

As a result, after being processed by our encoder module, the feature map may efficiently capture the highly condensed semantic representations and be more pixel-wise linearly independent.

### B. The Generalized Necessarily Distinct Vectors

Before introducing the concept of generalized necessarily distinct vectors, it is necessary to define the key concepts in linear algebra and matrix theory.

**Necessarily distinct vectors.** If we want to find a set of necessarily distinct vectors in a vector space, we must meet the following two requirements (see Fig. 6):

(a) This set of vectors must be linearly independent,

(b) The necessarily distinct vectors can represent any vector in the original feature map.

For requirement (a), according to the discussion above, we can conclude that when our encoder goes deeper, more convolutions in DenseNet Block and linear layers in Swin Transformer Stage are applied to the feature map, so after being processed by our encoder, the vectors in the feature map are linearly independent.

For requirement (b), we can say that the vectors coming from our guided self-attention have the ability to restore the original input image, that is, representing any vector in the original feature map. The reason is that the pixels in the matrix $\boldsymbol{B}$ represent a linear combination of kernel parameters and pixels from the previous feature map. Therefore, as our encoder goes deeper, the combination will become more complex, but will still be built of pixels from the previous feature map and parameters from the previous kernels. While dealing with restoring tasks, we can use alternative loss functions and back-propagation on the deconvolution to rebuild the low-level feature map. Therefore, the necessarily distinct vectors can represent any vector in the original feature map.

**Generalized necessarily distinct vectors.** As we discussed above, we can conclude that the vectors coming from the guided self-attention module are linearly independent and capable of restoring the lower-level feature map in the desired way. According to the definition of necessarily distinct vectors, all the discussed vectors must be in the same vector space [43]. Because those feature maps with different levels do not exist in the same vector space, the vectors coming from the guided self-attention module can be named "generalized necessarily distinct vectors", which means the vectors do not exist in the original image but can maintain the ability of the necessarily distinct vectors.

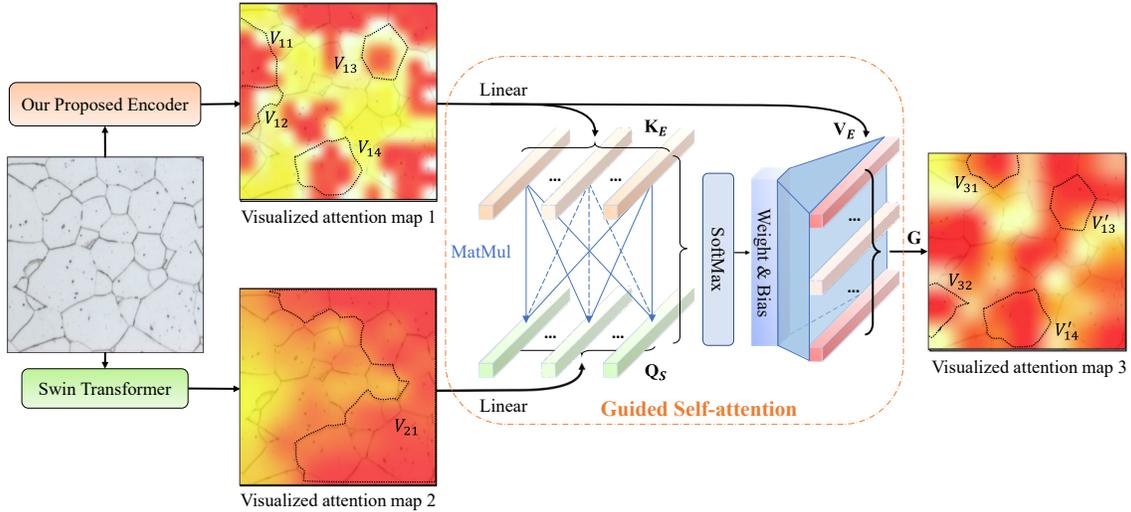

Fig. 7. Mechanism of our proposed Guided Self-attention module.

## C. Guided Self-Attention

We introduce a guided self-attention module designed to guide the model to find the generalized necessarily distinct vectors. These vectors facilitate the establishment of further long-distance connections based on the extensive local feature information processed through convolution. By incorporating our guided self-attention module, the model can explore deeper into the connections within the feature map, extracting more profound information. The conventional self-attention mechanism could be described as mapping a query and a set of key-value pairs to an output, where the query, keys, and values are all vectors linearly mapped by the same feature map. In this mechanism, extensive global interactions (i.e., cosine similarities between each query vector and all vectors in the keys) are extracted. However, this approach might direct the model's attention excessively towards long-range connections across the entire feature map, potentially diverting focus from relevant local features vital for the downstream task. As we can see from Fig. 7, the feature map processed by conventional self-attention mechanism (see the $V_{21}$ in the Visualized attention map 2) struggles to identify specific individual grains (e.g. $V_{13}$ and $V_{14}$ in Visualized attention map 1 of Fig. 7). Moreover, the Swin Transformer only depends on two linear layers of self-attention in each Swin Transformer Stage to apply dynamic weights and biases to each pixel, which may be inefficient for enhancing the pixel-wise linear independence when the training data is limited. This is evident in the accuracy comparison provided in Table I, where the Swin Transformer demonstrates lower accuracy in contrast to DenseNet. In contrast, each DenseNet Block in DenseNet121 comprises 12, 24, 48, and 32 convolution layers, which is more advantageous for enhancing pixel-wise linear independence in the feature map. Thus, as seen in Fig. 7, we apply our guided self-attention module to introduce key-value pairs (represented by matrices $\boldsymbol{K_E}$ and $\boldsymbol{V_E}$) from the new feature map processed by our encoder module (see Visualized attention map 1). Additionally, it introduces the query matrix (represented by matrix $\boldsymbol{Q_S}$) from the feature map processed by the conventional self-attention (see Visualized attention map 2). The dot production between $\boldsymbol{K_E}$ and $\boldsymbol{Q_S}$ allows the rich local information in $\boldsymbol{K_E}$ to guide $\boldsymbol{Q_S}$ to find the generalized necessarily distinct vectors, which can retain intricate relational connections and rich local feature information.

As shown in Fig. 7, "Visualized attention map 1", "Visualized attention map 2" and "Visualized attention map 3" are obtained by our proposed encoder module, the Swin Transformer, and our guided self-attention module, respectively. As we can see from Visualized feature map 1, although our proposed encoder can help the model merge the global information and local information more effectively (e.g., $V_{11}$ and $V_{12}$ are individual steel grains but having interaction with each other at the meanwhile), there still exist steel grains which are too far apart to interact with one another (i.e., $V_{13}$ and $V_{14}$). On the other hand, Swin Transformer can exploit global interaction better (i.e., the red part in the visualized attention map 2 of Fig. 7), but is not able to locate appropriate individual grains, which can represent the average grain size for grain size grading. Therefore, we introduce matrices $\boldsymbol{K_E}$ and $\boldsymbol{V_E}$ from the feature map processed by our proposed encoder (see Visualized attention map 1 in Fig. 7) via the linear layer, as well as linear mapping matrix $\boldsymbol{Q_S}$ from the feature map processed by Swin Transformer (see Visualized attention map 2 in Fig. 7). As we discussed above, $\boldsymbol{K_E}$ and $\boldsymbol{V_E}$ are more linearly independent and contain more local feature information than $\boldsymbol{Q_S}$, so we apply dot-product attention between $\boldsymbol{K_E}$ and $\boldsymbol{Q_S}$. Specifically, with the help of the dot-product attention, the cosine similarity between each vector in $\boldsymbol{K_E}$ and each vector in $\boldsymbol{Q_S}$ will be holistically calculated, enabling GSNet to concurrently use the benefits of matrices $\boldsymbol{K_E}$ and $\boldsymbol{Q_S}$. Then, we apply a softmax function on these cosine similarities to obtain the weights for $\boldsymbol{V_E}$. The formulation of the guided self-attention can be described as follows:

$$\boldsymbol{G} = softmax\left(\frac{\boldsymbol{Q_S}\boldsymbol{K_E^T}}{\sqrt{d_k}} + B\right)\boldsymbol{V_E} \qquad (7)$$

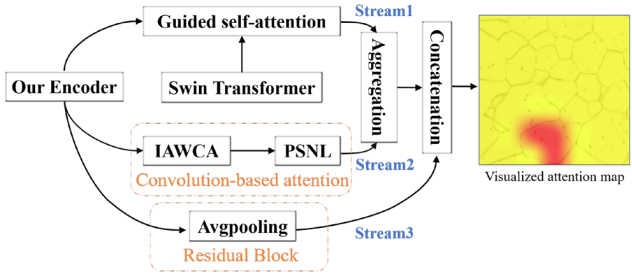

Fig. 8. Structure of triple-stream merging module.

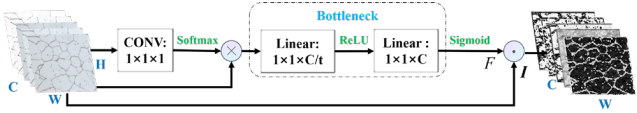

Fig. 9. Structure of IAWCA.

where $d_k$ indicates the dimension of queries and keys, and $B$ is a relative position bias [7]. As a result, with the help of our guided self-attention, GSNet can merge the local information and the global information more effectively, and produce the generalized necessarily distinct vectors (see the vectors $\boldsymbol{G}$ in Fig. 7) for grain size grading. The generalized necessarily distinct vector $\boldsymbol{G}$ can not only learn the interactions between distant grains (see $V'_{13}$ and $V'_{14}$ in the Visualized attention map 3 of Fig. 7), but also locate appropriate individual grains that can represent the average grain size for grain size grading (see $V_{31}$ and $V_{32}$ in the Visualized attention map 3 of Fig. 7).

In conclusion, the addition of the guided self-attention module can help the Swin Transformer find out the generalized necessarily distinct vectors, which can retain intricate relational connections and rich local feature information. Furthermore, the classification accuracy of our GSNet on the steel grain size datasets has increased by 0.6% after introducing the proposed guided self-attention.

### D. Triple-stream Merging Module

To enhance the model's generalization capability, we proposed the triple-stream merging module, consisting of three separate feature-processing streams, each representing diverse attention mechanisms and connection strategies. Through the incorporation of this module, the model can accurately identify a single grain that serves as a representative average size for steel grains (as depicted in the Visualized Attention Map in Fig. 8). As seen in Fig. 8, Stream1 represents the self-attention-based mechanism, which effectively merges local and global information through our proposed guided self-attention module. Stream2 signifies the convolution-based attention mechanism processed by IAWCA and PSNL (Patch-level Second-order Non-Local) modules [44], adept at extracting channel-wise information and global information, respectively. Stream3 denotes the feature map reusing mechanism processed by the residual block, capable of addressing the issue of vanishing gradient by introducing the skipping connection mechanism [28]. Specifically, we aggregate the feature maps from Stream 1 and Stream 2 through dot production, then concatenate the feature map from Stream 3 to create a more comprehensive merged semantic representation.

### 1) IAWCA.
In addition to exploiting local and global features, we have also proposed a channel-wise attention module, namely IAWCA (see Fig. 9), in our model. AWCA (Adaptive Weighted Channel Attention) [44] was proposed to reallocate channel-wise feature responses via integrating correlations between channels.

Due to the mechanism of Conv1x1, one output pixel can only be mapped from one vector via Conv1×1 in the bottleneck of AWCA, which may lose some important mapping relations between feature maps. On the contrary, in IAWCA, the linear layer in the bottleneck [45] can map one output pixel from all pixels in the input feature map, allowing IAWCA to preserve some important mapping relations between feature maps. Therefore, we replace the Conv 1×1 layer with the Linear layer (see Fig. 9) in the bottleneck of the AWCA module to fully exploit the local features. In the IAWCA module, denoting the input feature map as $\boldsymbol{I}=[i_1, i_2, ..., i_c, ..., i_C]$ containing $C$ feature maps with the size of $H \times W$, we exploit one convolutional layer to learn the adaptive weighted matrix $\boldsymbol{Y} \in R^{1 \times H \times W}$ and reshape $\boldsymbol{Y}$ to $R^{(H \times W) \times 1}$. Then we apply a Softmax layer to normalize $\boldsymbol{Y}$ and multiply $\boldsymbol{I}$ with $\boldsymbol{Y}$ before the bottleneck. The process is described as:

$$F = \delta(W_2(\sigma(W_1(\boldsymbol{I} \cdot \boldsymbol{Y})))),  \qquad (8)$$

where $W_1$ and $W_2$ are the weight sets of two fully connection layers, $\delta(\cdot)$ and $\sigma(\cdot)$ denote the sigmoid and ReLU activation functions, respectively. Then we assign channel attention map $F$ to rescale the input $\boldsymbol{I}$. With the IAWCA module, our model can efficiently adapt weights across channels, and the final accuracy can be improved by 0.4%.

### 2) PSNL.
As shown in Fig. 8, Stream1 can extract a lot of global information via the Swin Transformer. To make the feature maps processed by Stream1 and Stream2 more compatible, we introduce the PSNL into Stream2. PSNL is a convolution-based attention mechanism that is capable of capturing long-range dependencies across the entire image, which is similar to the shift window scheme in the Swin Transformer. Upon the application of PSNL to Stream2, the resultant feature map from Stream2 becomes enriched with extensive global information. This enrichment enhances the congruence and interoperability between the feature map of Stream2 and that from Stream1.

## IV. EXPERIMENTS

### A. Datasets and Evaluation Metrics

### 1) Dataset:
The steel grain size dataset contains 3,599 images with 14 grain size levels. In general, there is no precise threshold for the volume of data we may define as small. This concept is not autonomous, and is tied to model size and task complexity [46]. In the realm of steel, 300-500 images in one category can be defined as small [47], [48]. The steel grain size dataset may be regarded as a small dataset since there are 300-400 images for each level. These images are acquired from steel mills and have a resolution of 1376×1104 pixels. To address the overfitting problem, we crop the image into four 640×640 images along the four corners and consider six widely used data augmentations, namely Gaussian Blur, Random Grid Shuffle, Random Horizontal Flip, Random Vertical Flip, Random Rotation, and Coarse Dropout. Among the above six ways of augmentation, only the Random Horizontal Flip and Random Vertical Flip can improve classification accuracy by 0.4% and 0.6%, respectively. As a result, we apply Horizontal Flip to images with odd numbers and Vertical Flip to images with even numbers. After augmentation, we delete the duplicate pictures from different levels, then get 23,176 for training and 5,616 for validation. As for normalization, we mainly consider Batch Normalization (BN) [49] and Layer Normalization (LN) [50]. Since the channel-wise computation mechanism of LN is beneficial to the hierarchical architecture in Swin Transformer stages and Dense blocks, we apply the LN layer before the final classification layer to compute the mean and variance used for normalization, and the performance is better than BN by 0.5%. To regularize the model, we employ weight decay, which is usually applied to larger models and smaller datasets.

*2) Evaluation Metrics:* We adopt accuracy, mean average precision (mAP), recall rate, F1-score, explained variance score (EVS), mean squared error (MSE), and $R^2$ score to evaluate the performance of different models.

**Accuracy:** We have taken the $\pm 0.5$ level bias into consideration when evaluating the accuracy of models:

$$\begin{cases} acc_0 = num_0 \ / \ num \\ acc_{\pm 0.5} = num_{\pm 0.5} \ / \ num \\ acc = \alpha \times acc_0 + \beta \times acc_{\pm 0.5} \end{cases}, \quad (9)$$

where $acc_0$ denotes the accuracy that prediction equals to the ground truth, $num_0$ denotes the number of images predicted correctly, and $num$ denotes the total number of images. $acc_{\pm 0.5}$ denotes the accuracy that prediction equals to the ground-truth $\pm 0.5$, $num_{\pm 0.5}$ denotes the number of predictions with $\pm 0.5$ level bias, and $acc$ denotes the final classification accuracy on the steel grain size dataset, we take $\alpha = 0.4, \beta = 0.6$.

**Mean Average Precision (mAP):** We take the ground truth level as the true sample, the other levels as the false samples, the correct prediction as the positive sample, and the incorrect prediction as the negative sample. In this way, the mAP can be defined as follows:

$$mAP = \frac{1}{n} \sum_{i=1}^{n} P_i \ , \quad (10)$$

with

$$P_i = \frac{TP_i}{TP_i + FP_i} \ , \quad (11)$$

where $P_i$ denotes precision of the samples from the $i$-th grain size level, $TP_i$ denotes the $i$-th Ture Positive sample, $FP_i$ denotes the $i$-th False Positive sample, and $n$ denotes the number of grain size levels.

**Recall rate:**

$$Re = \frac{TP}{TP + FN}, \quad (12)$$

where $Re$ denotes the recall rate of all the samples, and $FN$ denotes False Negative samples. More true samples will be predicted positive when the recall rate is higher.

**F1-score:**

$$P = \frac{TP}{TP + FP} \ , F1 = \frac{2 \times P \times Re}{P + Re} \ , \quad (13)$$

where $P$ denotes precision of all the samples. Higher precision and recall rate lead to a higher F1-score.

**Explained variance score (EVS):**

$$EVS = 1 - \frac{Var\{y - \hat{y}\}}{Var\{y\}}, \quad (14)$$

where $y$ denotes the ground truth, $\hat{y}$ denotes the prediction, $Var\{y - \hat{y}\}$ denotes the variance of $\{y - \hat{y}\}$, and $Var\{y\}$ denotes the variance of $y$. When EVS is higher, the distribution of predictions is closer to that of ground truth.

**Mean squared error (MSE):**

$$MSE = \frac{1}{m} \sum_{j=1}^{m} (y_j - \hat{y}_j)^2 \ , \quad (15)$$

where $m$ denotes the number of samples, $y_j$ denotes the ground truth of the $j$-th sample, and $\hat{y}_j$ denotes the prediction of the $j$-th sample. When MSE is lower, the distribution of prediction is closer to that of ground truth.

**$R^2$ score:**

$$R^2 = 1 - \frac{MSE}{Var\{y\}}, \quad (16)$$

where $y$ denotes the ground truth, $MSE$ denotes mean squared error, and $Var\{y\}$ denotes the variance of $y$. When $R^2$ is closer to 1, the prediction is closer to the ground truth.

*B. Implementation details*

The proposed network is implemented based on the PyTorch frame [51] and trained for 63 epochs with a batch size of 16 using the SGD optimizer [52]. The weight decay and momentum are, respectively, set to 0.0001 and 0.9. We use the softmax cross-entropy loss function for multi-class classification. The learning rate is initially set to 0.001 and uses a polynomial decay with a power of 0.9. It takes 16 hours to train the proposed network on an NVIDIA PCIe A100 GPU.

*C. Performance*

*1) Comparisons with SOTAs:* As shown in Table II, we compare our model with several SOTA classification methods as well as some classical methods for classification, including SwinV2, Swin Transformer, Mae [53], Vit, DenseNet, ResNet, ResNeXt, Inception, and EfficientNet. The advantage of our GSNet lies in its low requirement for training data. To demonstrate this, we randomly selected 14 classes from ImageNet-1K, creating a much smaller dataset, ImageNet-14. Additionally, we chose CIFAR-10 as the second comparison dataset rather than CIFAR-100 due to its smaller number of categories, which is more comparable to the steel grain size dataset. We have evaluated our methods not only on the steel grain size dataset but also on CIFAR-10 and ImageNet-14, showcasing the robustness and generalizability of GSNet across different datasets. For ease of comparison, we have compared all models with the same input image size of $224 \times 224$. As we can see from Table II, our GSNet has the highest accuracy, even compared with those Swin Transformer-based models with ImageNet-22K pre-training. Moreover, on the steel grain size dataset, our GSNet has the best performance in terms of metrics like mean average

TABLE II

Model performance on the steel grain size dataset, Cifar10, and ImageNet-14 (#Acc (1/2): The Top-1 accuracy of the model without or with loading the pre-trained model, mAP: mean average precision, #PA: parameter size, #E: the number of epochs at which the model reaches the top-1 accuracy without loading the pre-trained model, #F: FLOPs (trillion), #Re: recall rate, #F1: F1-score, EVS: explained variance score, MSE: mean squared error, #$R^2$: $R^2$ score, #224: the Top-1 accuracy of the model without loading the pre-trained model with the image sizes of 224x224)

| Models | Steel Grain size dataset (224x224) | | | | | | | | | Cifar10 | ImageNet-14 |
| | #Acc1↑ / #Acc2↑ / mAP↑ | #Pa↓ | #E↓ | #F↓ | #Re↑ | #F1↑ | EVS↑ | MSE↓ | #$R^2$↑ | #224↑ | #224↑ |
| --- | --- | --- | --- | --- | --- | --- | --- | --- | --- | --- | --- |
| Swin-T | 0.758 / 0.873 / 0.573 | 106M | 129 | 0.06 | 0.566 | 0.559 | 0.932 | 0.242 | 0.929 | 0.743 | 0.648 |
| Swin-S | 0.760 / 0.879 / 0.591 | 188M | 91 | 0.13 | 0.581 | 0.566 | 0.941 | 0.206 | 0.941 | 0.749 | 0.691 |
| Swin-B | 0.764 / 0.883 / 0.594 | 332M | 125 | 0.24 | 0.580 | 0.571 | 0.938 | 0.211 | 0.938 | 0.733 | 0.678 |
| ResNeXt50 | 0.755 / 0.876 / 0.658 | 88M | 57 | 0.07 | 0.578 | 0.557 | 0.943 | 0.197 | 0.942 | 0.812 | 0.774 |
| ResNeXt101 | 0.789 / 0.874 / 0.610 | 332M | 52 | 0.26 | 0.601 | 0.591 | 0.940 | 0.257 | 0.939 | 0.764 | 0.781 |
| ResNet50 | 0.766 / 0.866 / 0.579 | 90M | 18 | 0.06 | 0.575 | 0.566 | 0.926 | 0.253 | 0.926 | 0.770 | 0.754 |
| ResNet101 | 0.769 / 0.869 / 0.630 | 162M | 49 | 0.12 | 0.623 | 0.609 | 0.951 | 0.167 | 0.951 | 0.773 | 0.764 |
| ResNet152 | 0.785 / 0.864 / 0.654 | 222M | 26 | 0.19 | 0.620 | 0.594 | 0.937 | 0.216 | 0.937 | 0.771 | 0.741 |
| EfficientNet-b0 | 0.761 / 0.875 / 0.558 | 15M | 54 | 6e-3 | 0.563 | 0.553 | 0.928 | 0.247 | 0.928 | 0.861 | 0.768 |
| ViT-B | 0.827 / 0.897 / 0.667 | 327M | 51 | 0.27 | 0.651 | 0.646 | 0.949 | 0.175 | 0.949 | 0.693 | 0.591 |
| ViT-L | 0.818 / 0.896 / 0.653 | 1.13G | 56 | 0.94 | 0.641 | 0.638 | 0.949 | 0.174 | 0.949 | 0.726 | 0.601 |
| Inception_v3 | 0.842 / 0.867 / 0.710 | 83M | 42 | 0.05 | 0.692 | 0.688 | 0.965 | 0.119 | 0.965 | 0.754 | 0.788 |
| DenseNet121 | 0.854 / 0.865 / 0.709 | 27M | 33 | 0.05 | 0.701 | 0.699 | 0.965 | 0.118 | 0.965 | 0.837 | 0.817 |
| DenseNet161 | 0.863 / 0.873 / 0.723 | 102M | 42 | 0.13 | 0.705 | 0.701 | 0.968 | 0.108 | 0.968 | 0.848 | 0.832 |
| DenseNet201 | 0.862 / 0.870 / 0.727 | 70M | 66 | 0.07 | 0.713 | 0.712 | 0.967 | 0.111 | 0.967 | 0.837 | 0.839 |
| SwinV2-T | 0.755 / 0.882 / 0.604 | 107M | 52 | 0.06 | 0.606 | 0.594 | 0.949 | 0.176 | 0.948 | 0.736 | 0.705 |
| Mae (Encoder) | 0.851 / 0.885 / 0.751 | 330M | 223 | 0.27 | 0.739 | 0.737 | 0.972 | 0.096 | 0.972 | 0.752 | 0.699 |
| ConvNext_CLIP | 0.810 / 0.881 / 0.649 | 3.1G | 22 | 2.42 | 0.638 | 0.635 | 0.959 | 0.167 | 0.951 | 0.818 | 0.637 |
| CoAtNet-Nano | 0.867 / 0.885 / 0.729 | 56M | 45 | 0.04 | 0.714 | 0.712 | 0.969 | 0.107 | 0.969 | 0.803 | 0.810 |
| EVA-02 | 0.879 / 0.890 / 0.766 | 329M | 77 | 0.28 | 0.752 | 0.750 | 0.977 | 0.079 | 0.977 | 0.733 | 0.722 |
| MaxViT-B | 0.889 / 0.892 / 0.774 | 454M | 72 | 0.37 | 0.764 | 0.762 | 0.976 | 0.082 | 0.976 | 0.815 | 0.845 |
| Our Encoder | 0.893 / 0.898 / 0.785 | 193M | 181 | 0.22 | 0.753 | 0.751 | 0.972 | 0.107 | 0.963 | 0.842 | 0.834 |
| **Our GSNet** | **0.901 / 0.904 / 0.785** | 349M | 63 | 0.27 | **0.771** | **0.769** | **0.978** | **0.076** | **0.978** | **0.897** | **0.854** |

TABLE III

Performance of GSNet, texture classification-based, and previous grain size grading methods on the steel grain size dataset. (#Top-1 Acc: The Top-1 accuracy of model, #Params: parameter size, #Speed: the time it takes the model to infer one image (seconds/one pic), #Epoch: the number of epochs at which the model reaches the top-1 accuracy)

| Model | #Top-1 Acc | #Params | #Speed | #Epoch |
| --- | --- | --- | --- | --- |
| Wavelet | 0.781 | 686M | 11.8 | 120 |
| FIC | 0.845 | 37M | 8.7 | 229 |
| GSNet | **0.901** | 349M | 42.3 | 63 |

TABLE IV

Performance of GSNet and preprocessing-based methods on the steel grain size dataset. (Method: The method used for image preprocessing, Classifier: The method used for classification, #Top-1 Acc: The Top-1 accuracy of the model, #Speed: The time it takes the model to infer one image (seconds/one pic))

| Method | Classifier | #Top-1 ACC | #Speed |
| --- | --- | --- | --- |
| Canny | SVR | 0.560 | 0.01 |
| Otsu | SVR | 0.561 | 2.29 |
| FWLBP | GSNet | 0.844 | 5.27 |
| Fuzzy | GSNet | **0.999** | 14.56 |

precision (mAP), recall rate, F1-score, EVS, MSE, and $R^2$ score. Besides, when the pretraining model is not loaded, our encoder module can outperform Swin Transformer and DenseNet121 in terms of metrics like accuracy, mAP, recall rate, F1-score, EVS, MSE, and $R^2$ score on the steel grain size dataset, and has higher accuracy on Cifar10 and ImageNet-14 than Swin Transformer and DenseNet121, demonstrating the effectiveness of our encoder module combining transformers and convolutions.

*2) Comparisons with texture classification method*: The grain size determination is somewhat similar to texture classification, so we make comparisons between our GSNet and two classical texture classification methods as follows.

Wavelet [54] is a texture classification model based on convolutions and uses the convolution and pooling layers as a restricted form of spectral analysis. Its classification accuracy on the steel grain size dataset is 78.1%.

FWLBP [55] is a preprocessor that can extract the texture information of the image (see Fig. 10(b)). FWLBP samples the image using an enhanced form of the local binary pattern (LBP) across three different radii, followed by an indexing operation to assign fractal dimension (FD) weights to the collected samples. After being processed by FWLBP, the steel grain size dataset is fed into GSNet, and the Top-1 classification accuracy is 84.4%. As shown in Tables III and IV, the classification accuracies of Wavelet and FWLBP are both lower than 0.901 of our GSNet.

*3) Comparison with previous grain size grading methods*. Table IV shows three previous grain size grading methods, in which Gajalakshmi et al. [35] used Otsu and Canny edge recognition to extract features for support vector regression (SVR) to identify the average grain size in a metallic microstructure. We re-implement the method in [35], and the Top-1 classification accuracies on the steel grain size dataset based on Otsu and

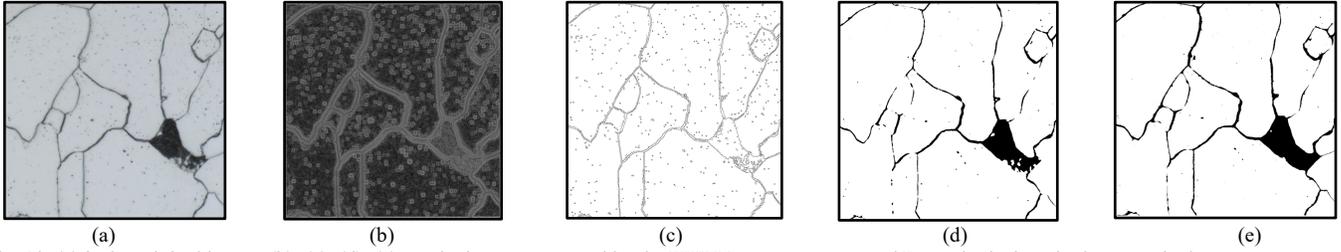

Fig. 10. (a) is the original image, (b), (c), (d), (e) are the image processed by the FWLBP, Canny, Otsu, and Fuzzy logical methods respectively.

TABLE V
PERFORMANCE OF FUZZY LOGIC METHOD WITH DIFFERENT CLASSIFIERS ON THE STEEL GRAIN SIZE DATASET. (**METHOD:** THE METHOD USED FOR IMAGE PREPROCESSING, **CLASSIFIER:** THE METHOD USED FOR CLASSIFICATION, **#TOP-1 ACC:** THE TOP-1 ACCURACY OF THE MODEL, **#EPOCH:** THE NUMBER OF EPOCHS AT WHICH THE MODEL REACHES THE TOP-1 ACCURACY)

| Method | Classifier | #TOP-1 ACC | #Epoch |
|--------|-----------|-----------|--------|
| Fuzzy | DenseNet121 | 0.9995 | 106 |
| Fuzzy | Swin-T | 0.9995 | 64 |
| Fuzzy | GSNet | **0.9996** | 46 |

TABLE VI
SETTINGS OF ABLATION EXPERIMENTS FOR GSNET (#ENCODER: ENCODER MODULE, #TRIPLE: TRIPLE-STREAM MERGING STRATEGY, #GUIDED: GUIDED SELF-ATTENTION MODULE, THE INPUT IMAGE SIZE IS 384×384)

| Model | #Encoder | #Triple | #Guided | IAWCA | #Acc |
|-------|----------|---------|---------|-------|------|
| GSNet-0 | ✓ | | ✓ | ✓ | 0.897 |
| GSNet-1 | ✓ | ✓ | | ✓ | 0.898 |
| GSNet-2 | ✓ | ✓ | ✓ | | 0.900 |
| GSNet-3 | ✓ | ✓ | ✓ | ✓ | 0.904 |

TABLE VII
MODEL PERFORMANCE ON THE STEEL GRAIN SIZE DATASET, CIFAR10, AND IMAGENET-14. (#GUIDED: THE GUIDED SELF-ATTENTION MODULE, #TRIPLE: THE TRIPLE-STREAM MERGING MODULE, #IAWCA: THE IAWCA MODULE.

| Model | Steel Grain size dataset | Cifar10 | ImageNet-14 |
|-------|--------------------------|---------|-------------|
| Swin-T | 0.758 | 0.743 | 0.648 |
| Swin-T+#Guided | 0.841 | 0.813 | 0.780 |
| Swin-T+#Triple | 0.863 | 0.801 | 0.778 |
| Swin-T+#IAWCA | 0.858 | 0.754 | 0.746 |

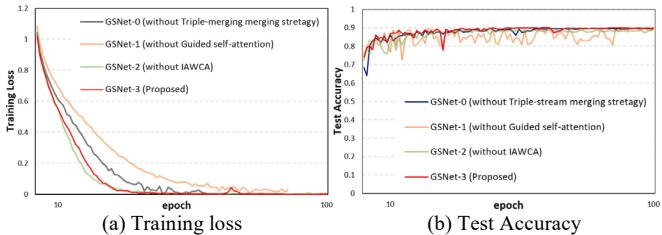

Fig. 11. Comparison of training loss convergence and test accuracy under various settings of the proposed architecture in GSNet.

Canny are 56.1% and 56.0% respectively.

Lee et al. [38] proposed convolution-based FIC (Fast Image Classification) for Grain size determination. On the steel grain size dataset, although the FIC is faster than our GSNet when inferring one single image, it takes 229 epochs to reach the Top-1 accuracy, while GSNet only requires 63 epochs (see Table III). In addition, the accuracy of FIC is 84.5%, which is lower than the 90.1% of GSNet.

Zhang et al. [34] proposed an image processing method to extract grain boundaries based on fuzzy logic. Although the pre-processing method could denoise the grain size picture efficiently, the paper [34] used the manual intercept method to grade the grain size level, which was inefficient and unreliable. Thus, we only apply this pre-processing method to the steel Grain size dataset. As shown in Fig. 10(e), the pre-processing method in [34] can effectively denoise the grain size image. We then apply our GSNet as a classifier on the processed Grain size dataset, and the Top-1 classification accuracy is improved to 99.9%. To confirm the correctness of the accuracy, we use Swin Transformer and DenseNet121 instead of GSNet as the classifier, and the results shown in Table V demonstrate the effectiveness of the fuzzy logic method. However, as shown in Table IV, the fuzzy logic method needs 14.56 seconds to process one picture, which is not conducive to practical production.

### D. Ablation Study

As discussed above, our encoder module can combine the convolutions and the transformers in an effective way. So, in this subsection, ablation experiments are conducted on the steel grain size dataset to verify the effectiveness of the proposed architecture, i.e., the guided self-attention module, the triple-stream merging module, and the IAWCA module. The experiment settings are shown in Table VI. The performance analysis is conducted for cases of whether to introduce our proposed modules or not. Additionally, we also visualize the training loss as well as the test accuracy of those ablation models at a resolution of 384×384 in Fig. 11. To substantiate the effectiveness of our modules, we have conducted several experiments integrating our modules with the Swin Transformer on the steel grain size dataset, Cifar10, and ImageNet-14. As shown in Table VII, the model performance improved significantly with the introduction of our module.

*1) Triple-stream merging strategy.* As Fig. 11(a) shows, GSNet-3 (red line) converges more quickly than GSNet-0 (gray line). Besides, the accuracy of GSNet-3 is higher than that of GSNet-0 by 0.7% after introducing the triple-stream merging strategy.

*2) Guided self-attention.* As Fig. 11(a) shows, GSNet-3 (red line) converges more quickly than GSNet-1 (orange line). Besides, as shown in Fig. 11(b), the GSNet-1 (orange line) is more fluctuating than that of the GSNet-3 (red line) and takes more time to reach the global optimal value, indicating that guided self-attention can assist the model in finding the optimal solution for classification more quickly and more stably, i.e., finding the generalized necessarily distinct vectors. Besides, the accuracy of GSNet-3 is higher than GSNet-1 by 0.6% after introducing the guided self-attention module.

*3)IAWCA module.* As Table VI shows, the accuracy of GSNet-3 is higher than that of GSNet-2 by 0.4% after introducing the IAWCA module. Thus, the IAWCA module may help the model capture more abundant information than the AWCA.

In contrast to the versions from GSNet-0 to GSNet-2, the proposed method (i.e., GSNet-3) can find the optimal solution for classification more quickly and stably. Consequentially, experimental results show that the proposed method paves the way for grain size determination.

*E. Failure Cases*

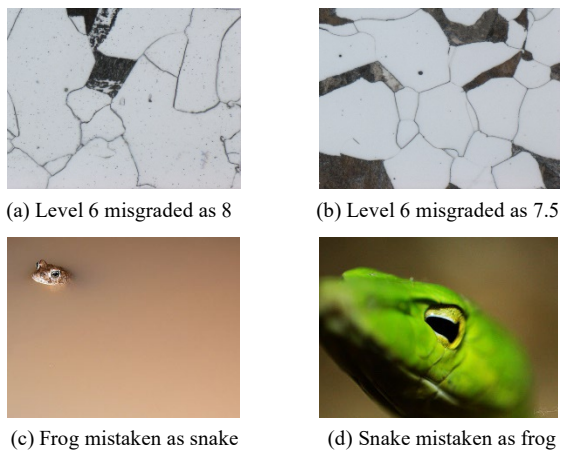

(a) Level 6 misgraded as 8     (b) Level 6 misgraded as 7.5

(c) Frog mistaken as snake     (d) Snake mistaken as frog

Fig. 12 Failure cases of GSNet

We observed that GSNet underperforms in the presence of abnormal grains, as indicated by the black regions in Fig. 12 (a) and (b), within the steel grain size dataset images. This observation suggests that the denoising capability of GSNet is limited. Additionally, in the ImageNet-14 dataset, GSNet demonstrates suboptimal performance when the target object is only partially visible (see Fig. 12(c) and (d), where the frog and snake share similar head and eye shapes). This finding implies that GSNet's key feature extraction ability has room for improvement.

## V. CONCLUSION

In this paper, we systematically take advantage of convolutions and Transformer to design a new family of models named GSNets. The key idea of the proposed method is to apply the guided self-attention module to assist the model in finding the generalized necessarily distinct vectors that can capture substantially condensed semantic representation for classification. Moreover, we propose an encoder module and triple-stream merging module to merge the local and global features more effectively, allowing the model to fully exploit pixel-wise linear independence. Extensive experiments show that the encoder module in GSNet demonstrates efficacy in integrating content-based global interactions with local features in the feature map.

Besides, the guided self-attention module can assist the model in finding the generalized necessarily distinct vectors more quickly and more stably. Furthermore, the triple-stream merging module can improve the generalization capability of the model by combining three feature-extracting strategies. Finally, GSNet can surpass state-of-the-art models on the small dataset. Although this paper focuses on classification, we believe our approach will be applied to broader applications such as object detection and semantic segmentation. We will leave them for future work.